\documentclass[conference]{IEEEtran}

\IEEEoverridecommandlockouts                    

\usepackage{graphics} 
\usepackage{float}
\usepackage{subfigure}

\usepackage{epsfig} 
\usepackage{mathptmx} 
\usepackage{times} 
\usepackage{amsmath} 
\usepackage{amssymb}  

\usepackage{cite}

\usepackage{autobreak}

\usepackage[flushleft]{threeparttable}
\usepackage{makecell,booktabs}
\usepackage{caption}

\usepackage{multirow}
\usepackage{array}
\usepackage{tabularx}
\usepackage{multirow, makecell}
\usepackage{balance}
\usepackage{epstopdf} 
\usepackage{svg}

\usepackage{color}

\usepackage{cuted}

\usepackage{dblfloatfix}    

\title{\LARGE \bf
Dual-arm Coordinated Manipulation for Object Twisting \\ with Human Intelligence
}

\author{
Weibang Bai,
Ningshan Zhang,
Baoru Huang,
Ziwei Wang$^*$,
Francesco Cursi, \\ 
Ya-Yen Tsai, 
Bo Xiao,
and Eric Yeatman,~\IEEEmembership{Fellow,~IEEE}

\thanks{
${*}$Coresponding authors.
This work was supported by EPSRC Programme Grant EP/P012779/1, UK Wellcome Trust GB-CHC-210183, European Commission Grants FETOPEN 829186, CONBOTS (ICT 871803), REHYB (ICT 871767), NIMA (FETOPEN 899626) and EPSRC Grant FAIR-SPACE EP/R026092/1.}
\thanks{W. Bai is with the Hamlyn Center and the Department of Computing, Imperial College London, London, SW7 2AZ, UK.}%
\thanks{N. Zhang 
was with University of Pennsylvania, Philadelphia, PA 19104, USA.}
\thanks{B. Huang, F. Cursi, Y. Tsai and B. Xiao are with the Hamlyn Center, Imperial College London, London, SW7 2AZ, UK.}%
\thanks{Z. Wang is with the Department of Biomedical Engineering, Imperial College London, London, SW7 2AZ, UK.}
\thanks{E. Yeatman is with the Department of Electrical and Electronic Engineering, Imperial College London, London, SW7 2AZ, UK.}%
}

\begin{document}

\maketitle
\thispagestyle{empty}
\pagestyle{empty}

\begin{abstract}

Robotic dual-arm twisting is a common but very challenging task in both industrial production
and daily services, 
as it often requires dexterous collaboration, large scale of end-effector rotating, and good adaptivity for object manipulation. Meanwhile, safety and efficiency are preliminary concerns for robotic dual-arm coordinated manipulation. Thus, the normally adopted fully automated task execution approaches based on environmental perception and motion planning techniques are still inadequate and problematic for the arduous twisting tasks.
To this end, this paper presents a novel strategy of the dual-arm coordinated control for twisting manipulation based on the combination of optimized motion planning for one arm and real time telecontrol with human intelligence for the other. 
The analysis and simulation results showed it can achieve collision and singularity free for dual arms with enhanced dexterity, safety, and efficiency.

\end{abstract}



\section{Introduction}
Collaborative manipulation with redundant dual-arm robotic systems have been widely adopted in various scenarios, 
due to the benefit of accuracy, efficiency and flexibility and the reducing the tedious workload and the demanded labor force for human society
\cite{tian2019configuration,smith2012dual}. 
%

Different coordination planning and control types and strategies have been developed for dual-arm systems.
%
Non-coordinated and coordinated manipulation, goal-coordinated and bimanual manipulation involving with non-contact and contact operations have been analyzed\cite{surdilovic2010compliance,park2016dual}.
Among the extensive methods for dual-arm coordination control, the leader and follower strategy \cite{luh1987constrained} is popular. It would set one arm as the main leader and the other as the follower during the motion-coordination control, and holonomic constrains can be deployed with the interacted object included within the closed chain. To implement the cooperative control technique, position mode and force mode were introduced and can be selected with flexibility \cite{hayati1989dual}. An anthropopathic coordinated control strategy based on real‐time approaching to a defined reference configuration for dual arms was also proposed to improve the coordination safety and efficiency \cite{bai2017novel}.

In terms of interacting with objects, different types of contact modeling were discussed to derive constraints for the coordinated manipulation \cite{smith2012dual,park2016dual}.
The rigid or non-rigid grasping modeling of the common object will affect the contact models\cite{smith2012dual,ni2019new}. Normally, the geometry and location information of rigid objects are known beforehand, and coordinated robotic grasping and manipulating can be achieved. But for elastic, flexible or deformable objects, the grasping state may change during the manipulation. As a result, it will cause further variation to the model chain, and the coordination motion planning may not work well with the derived cooperative control methods. In this case, the robot controller needs to be capable of estimating the current shape and state online for coordination. Shape estimation with extra sensory systems \cite{Cherubini2020}, local deformation model \cite{zhu2018dual}, internal force based impedance control strategy\cite{hsia2000internal}, and sliding mode control law \cite{al2007modeling} are developed for the flexible objects manipulation.

Teleoperation is also widely adopted for complex robotic manipulation tasks, as it provides better high-level policies and more intelligent decisions with enhanced adaptability and flexibility by virtue of using human intelligence. Traditional works mainly focused on the single-master-single-slave teleoperation framework \cite{Wang2020Adaptive,Wang2020Event,Wang2019Adaptive,CHEN2020Virtual}. Different from the aforementioned works, dual-arm configuration is more beneficial to facilitate complicated operation tasks. 
A tele-robotic framework for human-directed dual-arm manipulation was proposed in \cite{kruse2014sensor}, the operator uses hand motions to command the desired position for the object while the autonomous force controller maintains a stable grasp. In \cite{nicolis2018occlusion}, teleoperation of one of the dual-arm was applied to achieve occlusion-free visual servoing for shared autonomy control. In surgical robotic systems, multi-arms are normally controlled with teleoperation.
The real time environmental restrictions and coordination states are directly perceived and governed with human intelligence by the remote operator\cite{baidevelopment, bai2017modular}. 

\begin{figure*}[!t]
\centering
    \includegraphics[width = 0.75\textwidth, height = 0.22\textheight]{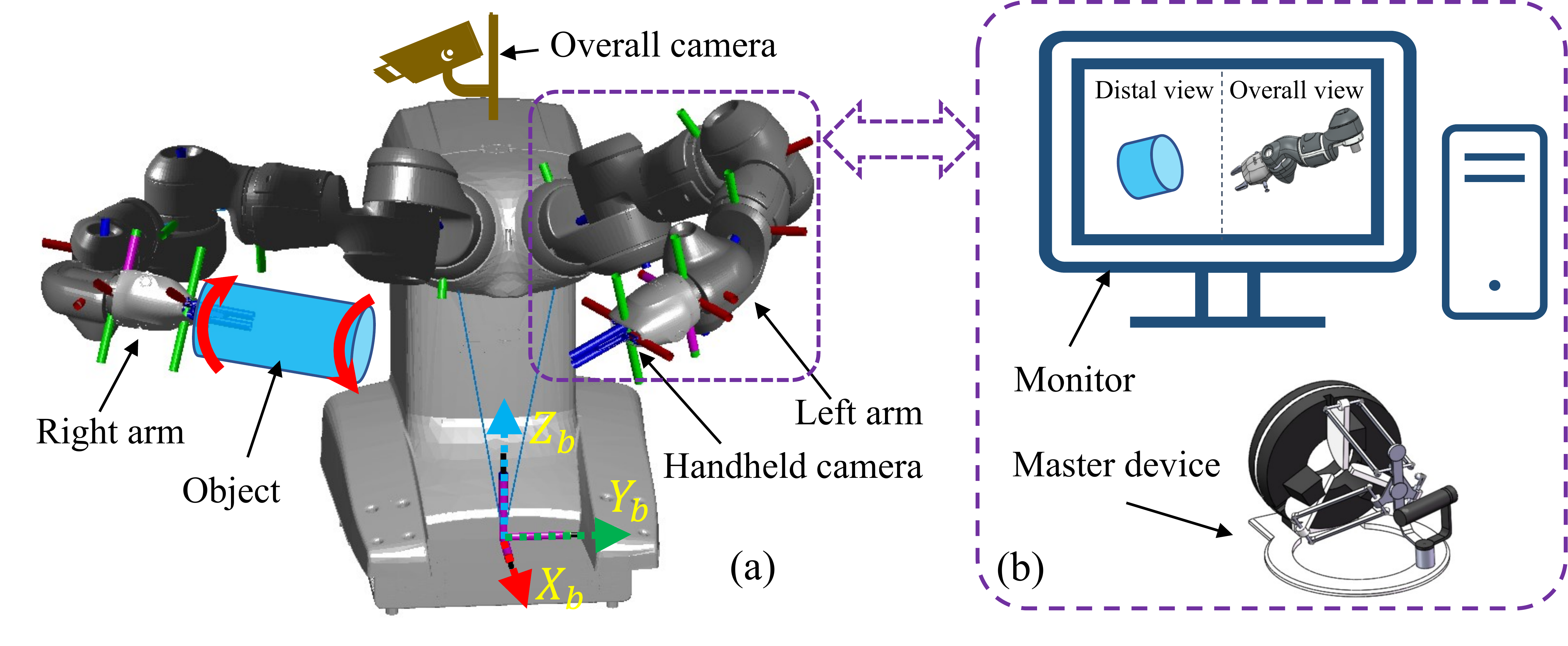}
    \caption{The systematic architecture of the dual-arm coordinated manipulation for object twisting with human intelligence. 
    (a)The dual-arm robot system with the right arm grasping an object for twisting and the left arm for teleoperation; (b)The master side setup for teleoperating the left arm with human intelligence for coordinated object twisting manipulation.}
\label{Fig.DualarmTwist}
\vspace{-5mm}
\end{figure*}%

To sum up, three main operation modes have been widely adopted for dual-arm manipulation: un-coordinated independent working sequences with collision avoidance; coordinated manipulation control based on different control strategies; and teleoperation control with human instructional inputs from remote side.

Dual-arm twisting manipulation, however, is a common but very difficult bimanual collaborative task with rigorous coordination constraints. It is more challenging as it is a kind of bimanual asymmetric non-congruent contact task \cite{sarkar1997dynamic, kruger2011dual, tanaka2005trajectory,vukobratovic2009dynamics } with many crucial restrictions.

In literature,
only a few have reported the issue of dual-arm twisting manipulation tasks. Some explorations are focused on industrial screw twisting tasks \cite{qiu2016hand}, which can be completed with one arm in many cases. 
In \cite{li2015twisting}, an algorithm took continuity and periodicity of joint angles into account to accomplish the tasks including twisting door handles and pulling doors open. 
In \cite{kim2011twisting}, a twisting door handle strategy which utilized the compliant motion control and the motion constraint for twisting with uncertain information was proposed, but it mainly focused on one arm twisting a fixed handle object. 
Another twisting technique was developed with joint limits and obstacles being minimized, and the motion of the manipulator was replaced by the in-hand manipulation\cite{takizawa2016implementation}.
But these twisting manipulation tasks are mostly completed around a fixed axis, which is not the general case in human daily life like twisting a towel or a bottle. Thus, further studies about dual-arm coordinated manipulation for object twisting are greatly needed.

However, the robotic dual-arm coordinated twisting manipulation is very challenging, especially when the spatial pose of the grasped object is not known or not accurate, such as
if the object is flexible or not grasped with precise pose as planned. But the online recognition and estimation of the shape and end pose difference of the object for coordination adjusting are not easy, especially when it is deformable while grasping and twisting. Moreover, with environmental geometry constraints like grasping from a narrow box or a stereoscopic warehouse, the dual-arm coordination workspace is restricted. This will cause more problems for grasping and twisting, like dual-arm grasping simultaneously may not be practicable.

In this paper, a dual-arm redundant collaborative robot YuMi (ABB Inc. Sweden)
\cite{ABBYumiManual} is used to complete the tasks of twisting manipulation.
To study and solve the problem for the dual-arm coordinated twisting manipulation, 
this paper presents a novel framework aiming to provide a general approach that can be suitable for the 
primitive and complicated situations with the aid of human intelligence. 
Firstly, one arm can be manipulated with principal motion planning, to grasp one end of the target object and adjust to provide feasible coordination. 
The other arm would be controlled by teleoperation to approach the object and grasp, despite the arbitrary end pose of the target object. Then, the twisting axis can be aligned by teleoperation for the dual-arm coordinated twisting.
The robotic systematic architecture of the framework can be illustrated in Fig.\ref{Fig.DualarmTwist}.

Meanwhile,
more procedures can be added into the framework to improve the safety, manipulability and efficiency of the dual-arm coordinated twisting. A geometry-based method that measures the distance between arms is derived to avoid the collision. A directional manipulability measure is applied for the pose adjustment and configuration optimization, and the minimization of the robotic motion variation are also implemented to provide better twisting manipulation.

\section{Problem Analysis}
\subsection{Coordinated object twisting constraints}

As stated previously, dual-arm twisting manipulation is a more challenging bimanual coordinated task with a lot of rigorous coordination constraints. 
\begin{itemize}
    \item Firstly, a large range of rotation for the both end effectors is needed, which should be performed with tolerable axial alignment but in the opposite circular directions in the meantime, to provide the torque for twisting execution. 
    \item Secondly, self-collision avoidance and kinematic singularity free are essential for the dual-arm cooperating. 
    \item Thirdly, if the operation is restricted with structural and environmental constraints like obstacles or narrow space, the manipulators will have to move with additional geometrical limitations. 
    \item Moreover, depending on the object properties, the twisted object may also affect the interaction and coordination, and in this case the interactive force, robotic dynamics and robot-environment interaction dynamics are even needed.  
    \item Furthermore, it will be more difficult especially when the object is flexible and deformable. The normally assigned common object frame is not fixed any more for the dual arms, the grasping state of one end of the object may not accurate and even change, and the resulted end pose of the object's other side for the second arm to hold and twist will be affected, thus the axis alignment may be inaccurate or even failed.
%
%
%

\end{itemize}
%

\subsection{Coordinated twisting task analysis}

In order to complete object twisting manipulation with dual-arm robots, we should define the problem in a solvable way. Dual-arm states, twisting objects, and twisting target motion need to be firstly clarified. 

\subsubsection{Dual-arm twisting description}
To make it more suitable for general cases in daily human life scenarios, environmental geometry constraints can be taken into account. Under normal circumstances, the object needs to be grasped from a narrow box or astereoscopic warehouse, and the dual-arm coordination workspace is restricted. Thus, for the dual-arm twisting operation, the first step of grasping the object may be only practicable using one arm, although it may be more convenient for the subsequent twisting when grasping and holding the object with two arms simultaneously. Therefore, as shown in Fig.\ref{fig.TaskStateDef}, we can define the main procedures for dual-arm coordinated twisting as follows:

\begin{figure}[!hb]
\centering
\subfigure[Grasping pose] {
    \label{fig.Grasppose}     
    \includegraphics[width=0.44\columnwidth]{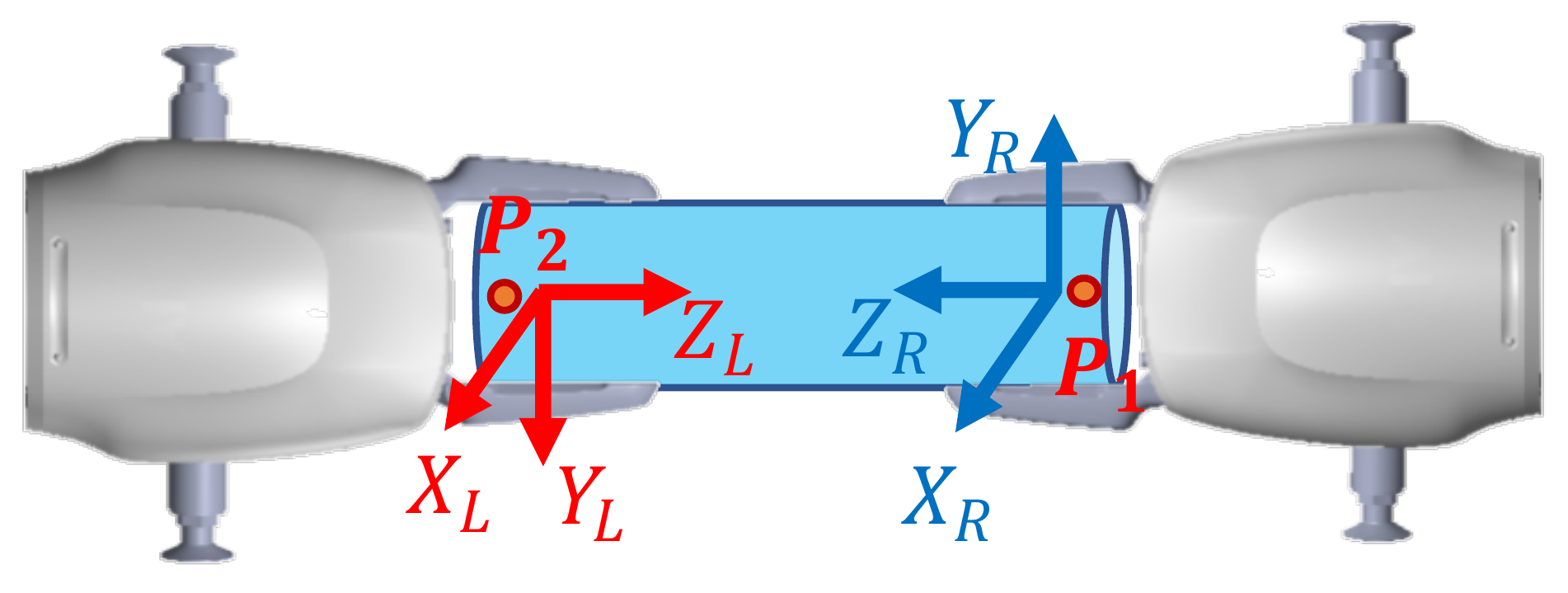}  
}     
\subfigure[Twisting pose] { 
    \label{fig.Twistpose}     
    \includegraphics[width=0.49\columnwidth]{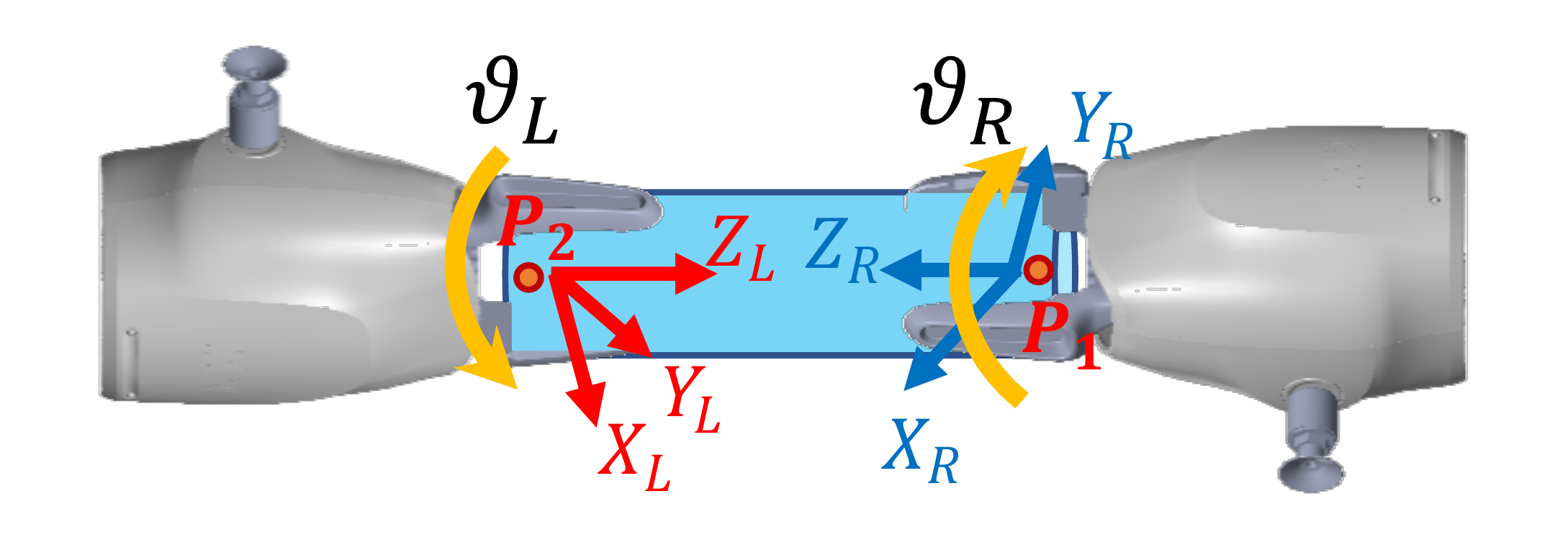}     
}

\caption{Task states definition.
(a) an ideal grasping state for the both end-effectors and the object, where no alignment error is indicated. (b) the twisting process with dual-arm rotated their end-effector in the opposite circular directions.}
\label{fig.TaskStateDef}
\vspace{-4mm}
\end{figure}

\begin{itemize}
\item Initial state: The dual-arm robot is starting from the initial state with standard configuration, where the pose of both end-effectors can be $X^{L}_{0}$ and $X^{R}_{0}$.

\item Grasping process: To address the general situation, one arm (the right arm) firstly moves its end-effector with frame $\{ RF_R \}$ to one end of the object with known spatial position $P_1(x_1, y_1, z_1)$ with respect to the robot base frame $\{RF_{b}\}$, i.e. $^{b}P^{R}_{e} \rightarrow{ ^{b}P_{1}}$, then hold and grasp the end $P_1$ along the object direction.
The grasped object will then be moved to a preparing location $P_{S}=(x_S, y_S, z_S)$,
where the other arm will move from its initial pose to the other end of the object $P_2(x_2,y_2,z_2)$ to hold the object together for the subsequent twisting manipulation.
The final ideal grasping pose for subsequent twisting can be
indicated in Fig.\ref{fig.Grasppose}, where the axes of both end effectors are aligned.

\item Twisting alignment: 
In real case, when the other arm (left arm) approaches 
the other end of the object $P_2$ at the preparing location $P_S$, its end pose $X^{L}$ should be adjusted to grasp and achieve the object twisting axis alignment. 
The alignment error $\delta$ for the dual-arm coordinated twisting manipulation should be within a tolerable misalignment $\delta_{tor}$, like $\delta_{tor} = 5^ \circ$. It can be defined as:
\begin{equation}
    \delta = <\overrightarrow{ z^R_e}, \ \overrightarrow{P_{1} P_{2}}>
\end{equation}
where $\overrightarrow{ z^R_e}$ is the direction vector of the right arm end-effector, and $\overrightarrow{ z^R_e}$ and $\overrightarrow{P_{1} P_{2}}$ should be in terms of the same frame.
If the stiffness of the elastic object is high, the object shape and its end position will not change much. The dual arms need to twist the object along the lengthwise axis and the error can be further defined as:
\begin{equation}
    \delta = <\overrightarrow{P^{L}_{e} P^{R}_{e}}, \ \overrightarrow{P_{1} P_{2}}>
\end{equation}
If the object is more flexible or deformable, the object itself will not provide large resistance torque when twisting or bending, and the dual arms can directly align the two end-effectors by adjusting the pose of the left arm. Then the error can be defined as:
\begin{equation}
    \delta = <\overrightarrow{^b z^R_e}, \ \overrightarrow{^b z^L_e}>
\end{equation}
where $\overrightarrow{^b z^R_e}$ and $\overrightarrow{^b z^L_e}$ are the direction vector of the two end-effectors in terms of $\{ RF_b \}$.

\item Twisting process: The two arms are twisting the object in the opposite circular direction, with the angle of $\vartheta_L$ and $\vartheta_R$ respectively, as shown in Fig.\ref{fig.Twistpose}. To simplify the problem, we can accept that the twisting process is done by reaching a threshold $\vartheta_T$ of the relative rotation angle $\vartheta_t$ which can be defined in a same frame like ${ \{RF_b\}}$ as:
\begin{equation}
    \vartheta_t = \lvert ^{b}\vartheta_L \ - \ ^{b}\vartheta_R \rvert
\end{equation}
Like we can set $\vartheta_T = 90 ^\circ$, and it can be achieved by rotating either one arm or both arms together.
\end{itemize}

In addition, the joint configuration of the dual arms can be optimized by finding the maximum manipulability for the twisting manipulation and the minimum of angle changes during the task.

\subsubsection{Twisting alignment problems}
\begin{figure}[htb]
\centering
 \includegraphics[width=0.63\columnwidth]{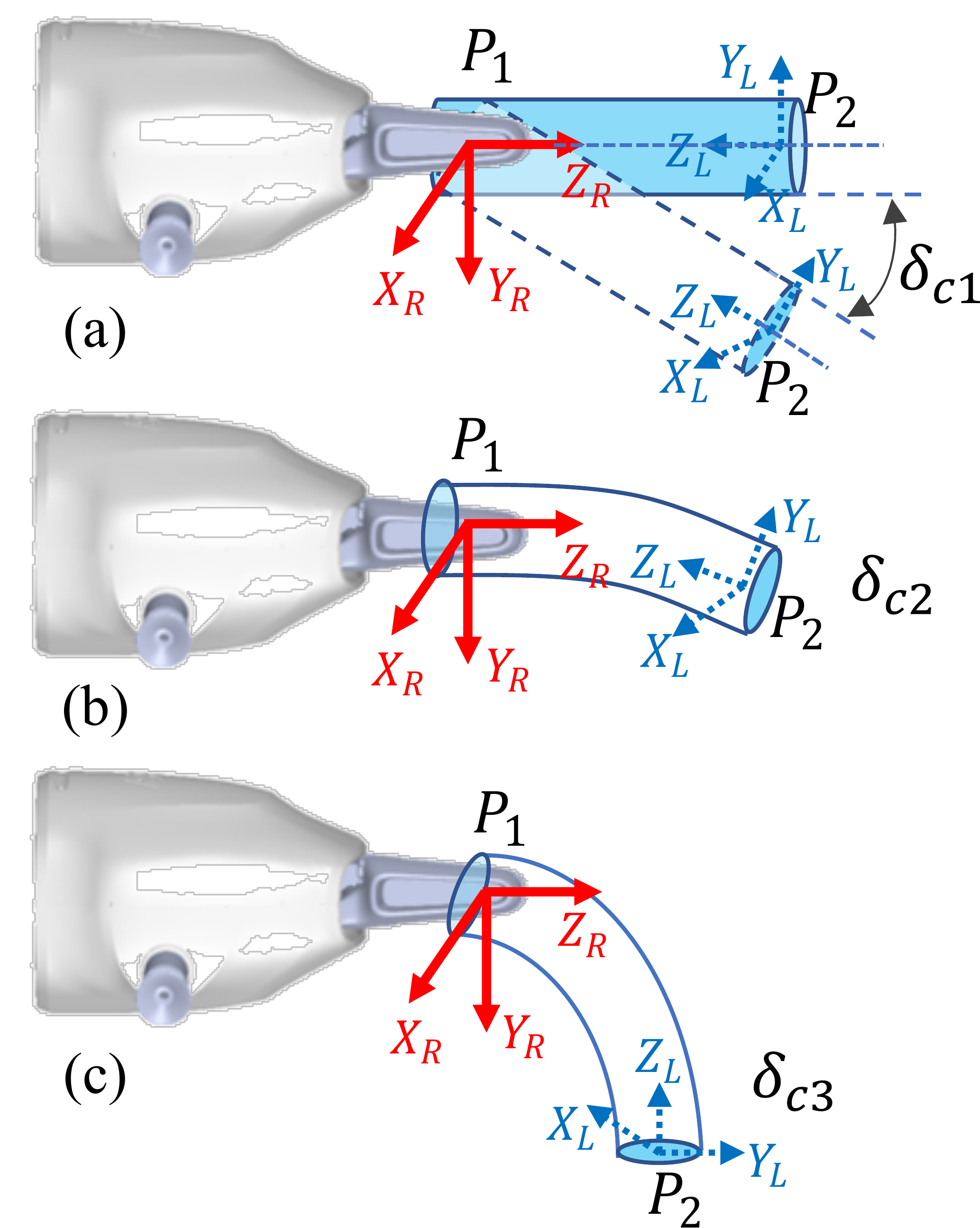}  

\caption{Illustration for twisting axis alignment problem caused by the grasped object, and the end-effectors' grasping coordinates are indicated along the both ends of the object. (a) For higher stiffness objects, the grasping pose for the first arm may be not accurately achieved as planned or changed for some reason. (b) For flexible objects, elastic bending angle may be generated on the other end. (c) For soft objects, large varied angle will happen on the other side.}
\label{fig.AlignmentProblem}
\vspace{-6mm}
\end{figure}

Normally the twisting object can be briefly simplified as a columnar bar, columnar or prismatic, as shown in Fig.\ref{Fig.DualarmTwist} and Fig.\ref{fig.TaskStateDef}. 
However, the twisting object can be both with high stiffness or more flexible, which would cause different problems for dual-arm autonomous twisting manipulation when completing the grasping process and twisting alignment as defined previously. When the object is with high stiffness, the first arm's grasping state may be not as accurate as planned or may change for some reason before the second arm's grasping and twisting, and a varied angle $\delta_{c1}$ will be generated, as shown in Fig.\ref{fig.AlignmentProblem}(a). When the object is flexible but the stiffness is not that low, the other end of the grasped object will bend due to the gravity with a varied angle $\delta_{c2}$, as shown in Fig.\ref{fig.AlignmentProblem}(b). When the object is flexible and even soft, like twisting a towel, the other end will be with large deformation, and the varied angle $\delta_{c3}$ will even be nearly $90^\circ$ when the first arm hold the object end horizontally, as shown in Fig.\ref{fig.AlignmentProblem}(c).

In this situation, the varied angle on the other end of the object $\delta_{c1}, \ \delta_{c2}, \ \delta_{c3}$ will greatly affect the alignment of the second arm to grasp and twist.
Twist alignment failure or inaccuracy will further make the dual-arm coordinated twisting failed. What is more, the varied angle can be dynamic and changing when the arm is moving. It is, therefore, not easy to model or detect it accurately in real time, 
which makes it a great challenging for the real time autonomous control of the second arm to approach and grasp the other end $P_2$ with accurate axis alignment and the further coordination modeling and control for the dual-arm twisting manipulation.


\subsection{Concept}

To address the challenging twisting alignment problems, real-time intelligent detection of the spatial pose of the other end of the object $P_2$ and proper grasping are needed for dual-arm coordination,
but these still remain to be problematic for the robotic autonomous manipulating as it needs accurate and fast sensing, real-time motion planning and dynamic control.

On the other hand, robotic teleoperation control with human intelligence can provide high-level real-time motion planning and intelligent control policies based on the on-site information feedback from the arm side.  
Thus the twisting alignment and the subsequent twisting manipulating process can be intelligently and efficiently tele-controlled by the human operator. The proposed new framework is shown in Fig.\ref{Fig.DualarmTwist}.

Meanwhile, as indicated in \cite{hayati1989dual}, if the object is not rigid and the control of internal forces is not cared, then the 
dual-arm coordination manipulation can be controlled in position mode. In object twisting tasks, the object is elastic, deformable or even soft, the interactive force and the resistance torque can be assumed as negligible, and the internal forces control is not necessary. So we can simplify the problem by implementing dual-arm coordination control in position mode, which is also beneficial to the robotic teleoperation control.
%

In addition, we still need to consider the coordination manipulation. The safety issues about self-collision, the redundant arm configuration optimization for improving task efficiency and enhancing the capacity of anti-interference as the opposite torsion will be exerted on each other, and the total motion minimizing of the dual-arm can be further addressed.

\section{Methods}

\subsection{Teleoperation architecture}
In the proposed framework, one arm (the right arm) is controlled with pre-defined motion planning, its end-effector will firstly approach the target object,
grasp the end of the object $P_1$, and move the object to $P_S$. While the other arm (the left arm) is controlled through teleoperation. With the overall and handheld cameras mounted in the slave side, the human teleoperator can make immediate decisions and plan appropriate motions for the other end-effector in real-time to approach the other end of the object $P_2$ and grasp. 

The master side motion commands is obtained from human operator controlling a haptic device Omega 7 (Force Dimension Inc.,Switzerland) in real time. The open and close of the end-effector can be linearly mapped to the gripper of the 7 DOF master device.

The general Cartesian space kinematic end-to-end mapping can be developed the master-slave teleoperation \cite{rezazadeh2019robotic}. With the robotic forward and inverse kinematics function analytically described as $f_{k}(\cdot)$ and $f_{k}^{-1}(\cdot)$, the mapping function $g\left(q_{m} \mapsto q_{s}\right)$ can be expressed as:
\begin{equation}
\setlength\abovedisplayskip{4pt}
\setlength\belowdisplayskip{4pt}
\begin{aligned}
X_s = g(X_m) = g\left(^m f_{k}\left(q_{m}\right)\right) 
\end{aligned}
\end{equation}
Where $X$ is the end pose, $q$ is the joint state, $m$ and $s$ represent the master and slave sides of the teleoperation. Then the target joint state of the slave robot arm would be:
\begin{equation}
\setlength\abovedisplayskip{4pt}
\setlength\belowdisplayskip{4pt}
\begin{aligned}
q_{s} = ^s f_{k}^{-1}\left(X_{s}\right) = \,^s f_{k}^{-1}\left(g\left(\,^m f_{k}\left(q_{m}\right)\right)\right)
\end{aligned}
\end{equation}

In joint space, the target joint value in each control loop can be further restricted with more constraints when teleoperation, such as the arm collision avoidace for dual-arm coordination, manipulability enhancement, joint configuration variation reduction and singularity free. 

\subsection{Self-collision detection}

For dual-arm manipulators, the obstacles are not only from the external environment but also can be from the other arm that is constantly moving as two arms are sharing the same workspace during the operation. In order to achieve successful coordination and perform safe collision-free motion, the collision detection and collision avoidance for the dual-arm manipulators are necessary.
\begin{figure}[htp]
\centering    
\subfigure[Line to Line] {
    \label{fig.collimod.a}     
    \includegraphics[width=0.43\columnwidth]{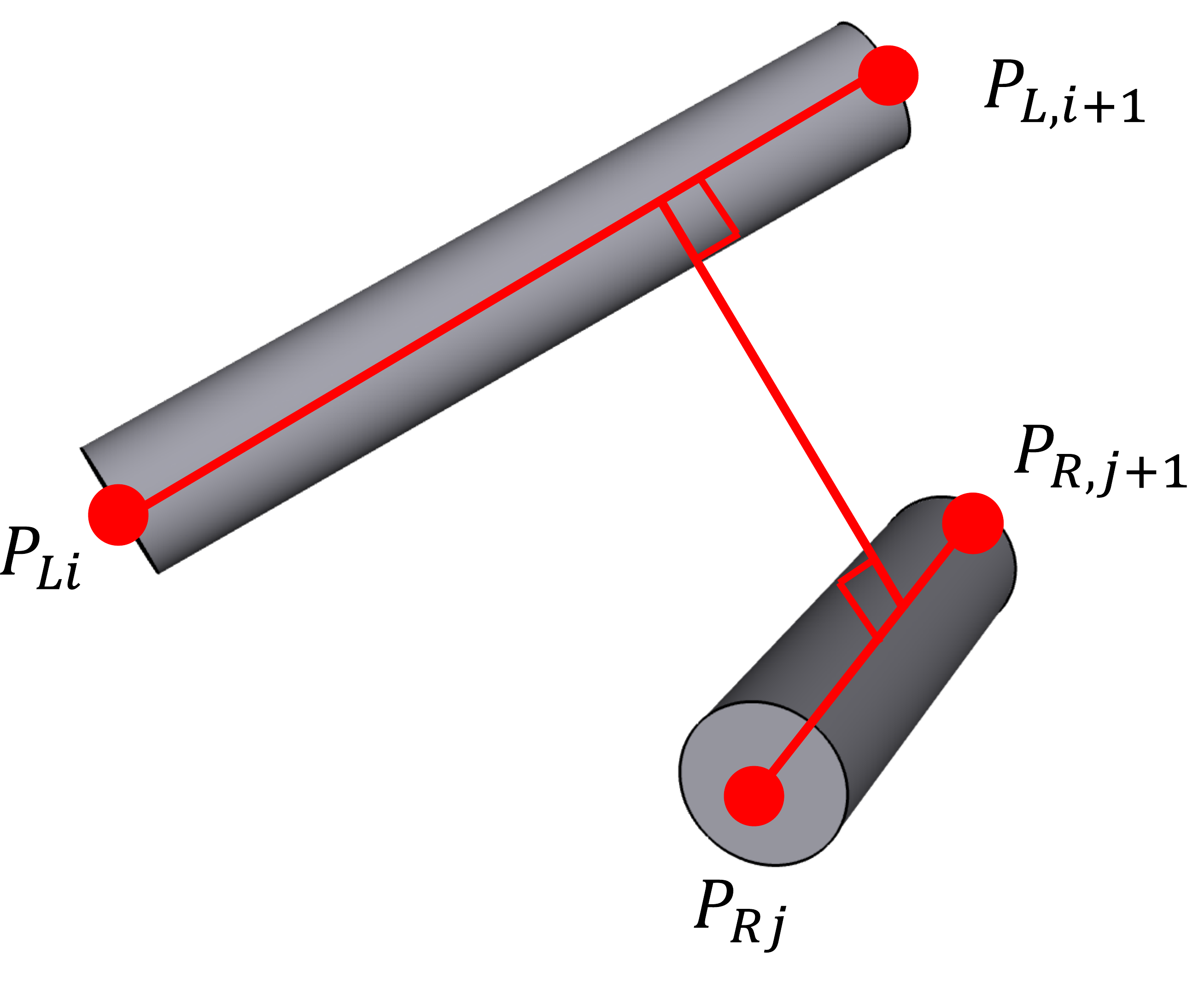}  
}     
\subfigure[Point to Line ] { 
    \label{fig.collimod.b}     
    \includegraphics[width=0.4\columnwidth]{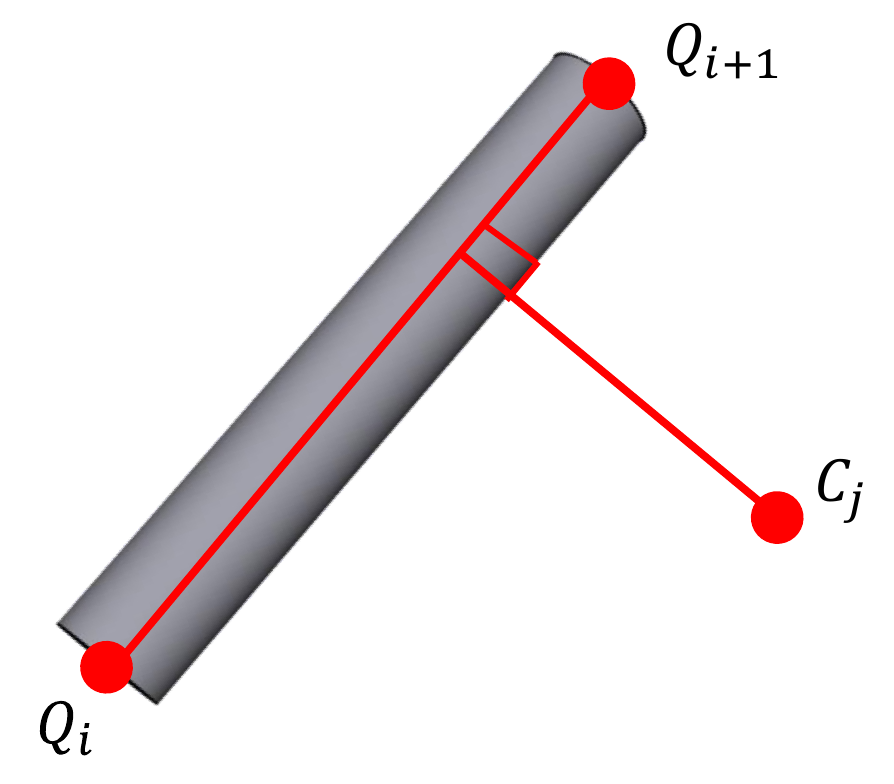}     
}    
\caption{Collision detection models. }     
\label{fig.collimod}
\vspace{-4mm}
\end{figure}

The minimal distance between two manipulators can be derived based on the basic models of distance between two lines and a point to a line, as shown iin Fig.\ref{fig.collimod}.
Fig.\ref{fig.collimod.a} shows the distance model of two arbitrary links of the two manipulators. $P_{Li}$ and $P_{L, i+1}$ represent the joint $i$ and joint $i+1$ of left arm, $P_{Rj}$ and $P_{R, j+1}$ represent the joint $j$ and joint $j+1$ of the right arm. The positions of these joints can be obtained from forward kinematic equations of each arm with the corresponding joint configurations.

Thus the spacial linear equations of two links could be obtained with analytic geometry representations:
\begin{equation}
\left\{\begin{array}{l}
P_{Li}\left(c_{Li}\right)=P_{L i}+c_{Li} \boldsymbol{n}_{Li} \quad (i = 1,2, \dots, N_L)\\[0.5ex]
P_{Rj}\left(c_{Rj}\right)=P_{R j}+c_{Rj} \boldsymbol{n}_{Rj} \quad (j = 1,2, \dots, N_R)
\end{array}\right.
\end{equation}
where $n_{Li}$ and $n_{Rj}$ are the direction vectors of $\overrightarrow{P_{Li}P_{L,i+1}}$ and $\overrightarrow{P_{Rj}P_{R,j+1}}$ respectively, $N_L$ and $N_R$ are the joint number of the two arms, and $c_{Li}$ and $c_{Rj}$ are the coefficients. 
The distance $d_{lij}$ between the two lines would be:
\begin{equation}
d_{lij}=\frac{\left(\boldsymbol{n}_{Li} \times \boldsymbol{n}_{Rj}\right) \cdot\left[P_{Li}\left(c_{Li}\right)-P_{Rj}\left(c_{Rj}\right)\right]}{\left|\boldsymbol{n}_{Li} \times \boldsymbol{n}_{Rj}\right|}
\end{equation}

Given the smallest collision distance threshold $d_{thr}$, if $d_{lij}\geq d_{thr}$, the collision would not happen between the two lines. But when $d_{lij}<d_{thr}$, $c_{Li}$  and $c_{Rj}$ should be calculated first:

\begin{equation}
\left\{\begin{aligned}
c_{Li}=& \frac{
\left(\boldsymbol{n}_{Li} \cdot \boldsymbol{n}_{Rj}\right)\left[\left(P_{L i}-P_{R j}\right) \cdot \boldsymbol{n}_{Rj}\right]
-
\left|\boldsymbol{n}_{Rj}\right|^{2} \left[\left(P_{R j}-P_{L i}\right) \cdot \boldsymbol{n}_{Li}\right]}
{\left|\boldsymbol{n}_{Li}\right|^{2} \left|\boldsymbol{n}_{Rj}\right|^{2}-\left(\boldsymbol{n}_{Li} \cdot \boldsymbol{n}_{Rj}\right)^{2}} 
\\
c_{Rj}=& \frac{
\left|\boldsymbol{n}_{Li}\right|^{2} \left[\left(P_{L i}-P_{R j}\right) \cdot \boldsymbol{n}_{Rj}\right]
-
\left(\boldsymbol{n}_{Li} \cdot \boldsymbol{n}_{Rj}\right)\left[\left(P_{R i}-P_{L j}\right) \cdot \boldsymbol{n}_{Li}\right]}{\left|\boldsymbol{n}_{Li}\right|^{2} \left|\boldsymbol{n}_{Rj}\right|^{2}-\left(\boldsymbol{n}_{Li} \cdot \boldsymbol{n}_{Rj}\right)^{2}}
\end{aligned}\right.
\end{equation}
If $c_{Li},\ c_{Rj}\in\left[0,\ 1\right]$, the minimum distance between the links is $d_{lij}$, and it is unsafe. If $c_{Li},\ c_{Rj}\notin\left[0,\ 1\right]$, the point that defines the minimum distance segment is located on the extension lines of the links, and further calculation of the distances between the lines and the joints on the other arm are needed. 
%
%
As shown in Fig.\ref{fig.collimod.b}, $Q_i$ and $Q_{i+1}$ represent the position of joint $i$ and joint $i+1$, $C_j$ represents one of the joints $j$ of the other arm. The line $Q_iQ_{i+1}$ is expressed by:
\begin{equation}
Q(b)=Q_{i}+b \textbf{\emph {s}}
\end{equation}
where $b$ is the coefficient and $\textbf{\emph{s}}$ is the direction vectors of $\overrightarrow{Q_iQ_{i+1}}$. The distance $d_p$ between the point to the line is:
\begin{equation}
d_{p}=\left|\overrightarrow{Q_{i} Q_{i+1}} \times \overrightarrow{Q_{i} C_{j}}\right| / \left| \overrightarrow{Q_{i} Q_{i+1}} \right|
\label{equ:dp}
\end{equation}

If $d_p\geq d_{thr}$, the 
collision would not occur. If $d_p<d_{thr}$, the foot of the perpendicular should be detected whether it is on the line or its extension by:
\begin{equation}
b=\left| \overrightarrow{Q_{i} Q_{i+1}} \cdot \overrightarrow{Q_{i} C_{j}} \right| / \left| \overrightarrow{Q_{i} C_{j}} \right|^{2}
\label{equ:b}
\end{equation}
If $b\in\left[0,\ 1\right]$, it is on the line, the minimal distance is $d_p$, and the collision occurs; otherwise, it is on the extension line and the minimum distance will be $\min \left\{ |Q_{i}C_j|,\ |Q_{i+1}C_j| \right\}$.

The final minimal distance $d_{min}$ between the two arms can be obtained by comparing and choosing from all the calculations above with $i = 1,2, \dots, N_L$ and $j = 1,2, \dots, N_R$.

\subsection{Directional manipulability evaluation}
Manipulability \cite{yoshikawa1985manipulability} of a manipulator is the index measuring its kinematic performance.
Higher manipulability is useful for robotic flexibility and adaptability, especially for the demanding twisting task.
Good joint configurations of grasping could benefit the twisting process. The directional manipulability \cite{yao2000task} can be used to indicate the manipulability in the specific motion direction:
\begin{equation}
M= 1 / k^{T}\left(J J^{T}\right)^{-1} k
\end{equation}
where $k$ is the directional unit vector of velocity, $J =[J_v \ J_\omega]^T$ is the Jacobian matrix of the manipulator.
For the dual-arm twisting task, 
the angular velocities of both arms $\omega_L$ and $\omega_R$ are needed to be considered during the twisting process. 
\begin{equation}
\omega_{i}=W_{i} k_{i}, \quad ( i = \{R,L\} ) 
\label{equ:wlr}
\end{equation}
where $R$ and $L$ stand for the right and left arm. $W_i$ represents the magnitude of the angular velocity. 

Based on the manipulator kinematics and dynamics theory, we have $\dot x=J \dot \theta$ and $\tau=J^T W$, where $W=[f^T \ n^T]^T$ is the wrench at the end-effector. Thus the velocity and force manipulability ellipsoids are orthogonal to each other, and the two ellipsoids have principal axes along the same direction with reciprocal magnitude. To enhance the anti-interference capacity of the torsion generated by the twisting motion, which means larger torque about the opposite axial directions can be provided for external needs, we can maximize the rotational force ellipsoid especially the magnitude of the moment term about the object twisting axis. 
As momentum can be exerted bidirectionally about an axis in real case, the target direction can be just assigned just along the twisting axis for our task.
The unit directional vectors can be specified in the end-effectors' own frame $\{ RF^L_e \}$ and $\{ RF^R_e \}$ as:
\begin{equation}
\left\{\begin{array}{l}
^{eL} k_L=\left[0,\ 0,\ 1\right]^T \\[0.25ex]
^{eR} k_R=\left[0,\ 0,\ 1\right]^T
\end{array}\right.
\label{equ:drectvect}
\end{equation}
With the directional manipulability index, we have:
\begin{equation}
M_{i}=1/{k_{i}}^{T} \left( J_{\omega, i} \ J_{\omega, i}^{T} \right)^{-1} k_{i}, \quad ( i = \{R,L\} )
\end{equation}

To maximize the specified directional manipulability, the reciprocals can be minimized and the fitness function can be:
\begin{equation}
\begin{array}{l}
\mathop{\arg \min}
f_{m}=\sum \frac{\beta_i}{M_{i}} 
\end{array}
\end{equation}
%
where $\beta_i$ is the weight coefficient that can be defined manually.

\subsection{Weighted configuration variation}
During the manipulation task, we can minimize the movement angle of each joint and further reduce the amplitude of the motion variation of the entire manipulators.
The change of each joint with different weights can be written as:
\begin{equation}
\left\{\begin{array}{l}
A_{L i}=\alpha_{i}\left|\theta_{L i}^{\text {final}}-\theta_{L i}^{\text {initial}}\right|, (i = 1,2, \dots, N_L) \\[0.75ex] 
A_{R j}=\alpha_{j}\left|\theta_{R j}^{\text {final}}-\theta_{R j}^{\text {initial}}\right|, (j = 1,2, \dots, N_R) 
\end{array}\right.
\end{equation}
where $A_{Li}$ and $A_{Rj}$ represent the weighted angle changes for both arms
from initial configuration to final target configuration. $\alpha_i$ and $\alpha_j$ are the weighting factors of the variation of each joint. The configuration weighting factors for both arms are set as: $\alpha_1=1,\ \alpha_2=\alpha_3=0.5,\ \alpha_4=\alpha_5=\alpha_6=\alpha_7=0.1$. In this case, less motion is allocated to the joints close to the base and more is allowed for the joints close to end-effectors, and the whole motion scale can be reduced.
The fitness function then can be expressed as:
\begin{equation}
\begin{array}{l}
\arg \min f_{a}=\sum_{i,j=1}^{7}\left(A_{L i}+A_{R j}\right) \\[1mm]
\text { s.t. }\left\{\begin{array}{l}
| J_{L}|\neq 0, \ | J_{R} \mid \neq 0 \\[1mm]
\theta_{R j} \in\left[\theta_{R j, \min }, \theta_{R j, \max }\right] \\[1mm]
\theta_{L i} \in\left[\theta_{L i, \min }, \theta_{L i, \max }\right]
\end{array}\right.
\end{array}
\label{equ:fa}
\end{equation}
where the Jacobian matrices of left and right arm $J_L$ and $J_R$ are used to avoid the singularities, and each joint value should be within its own joint limit.

\section{Simulation and Results}

\begin{figure}[htp] 
    \centering
    \includegraphics[width = 0.48\textwidth]{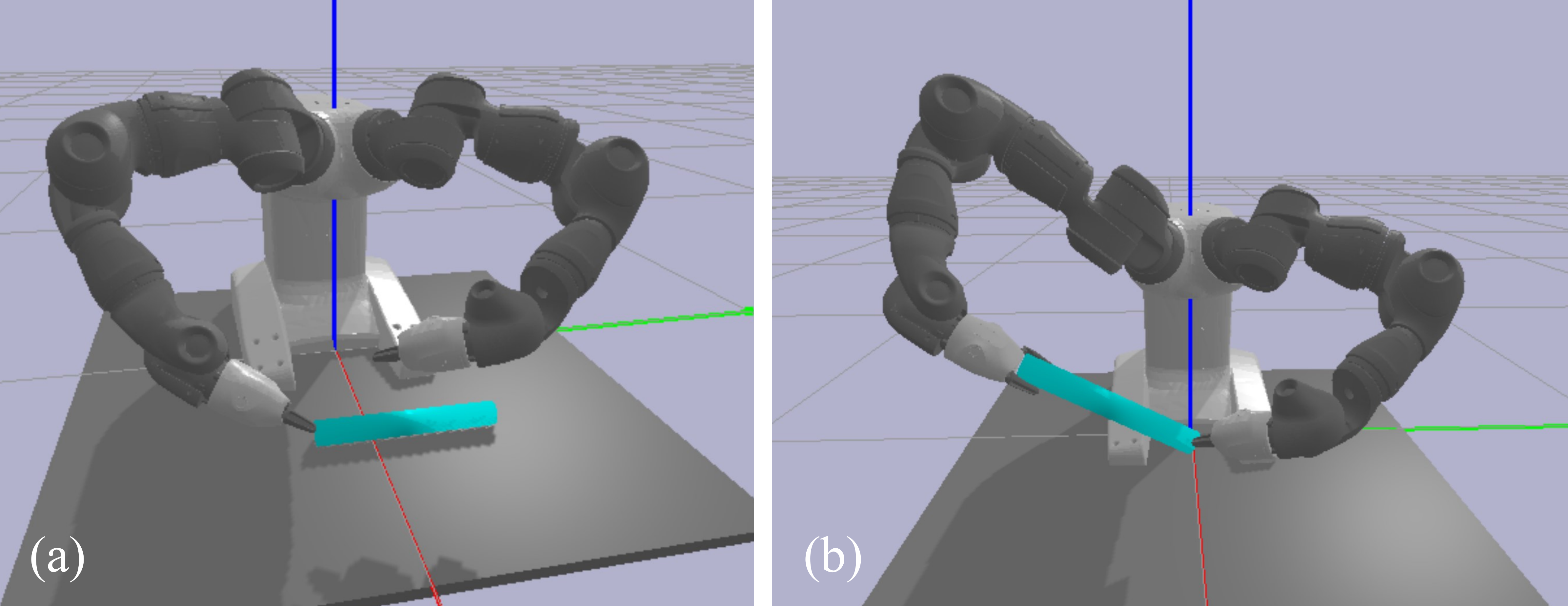}
    \caption{Simulation states: (a) Starting state: the right arm starts to grasp at one end of a cylindrical object which is flexible but with a certain stiffness. (b) Idea twisting alignment state: the right arm lifts the grasped end to a preparation location, the teleoperated left arm grasps the other end and adjusts its spatial pose by trying to keep the both ends in a same line to reconstruct the cylinder and achieve twisting alignment.}
    \label{fig.robosim}
    \vspace{-6mm}
\end{figure}

The proposed dual-arm coordinated manipulation method for the object twisting task is simulated with PyBullet. For the convenience of simulation, the sample flexible object is defined by a cylinder with a certain stiffness which is deformable, as indicated in the case of Fig.\ref{fig.AlignmentProblem}(c). During the simulation, the right arm will firstly move its end-effector to one end of the cylindrical object and grasp, then lift and move the grasped object to a preparation location. The other end of the object will fall due to gravity. Afterwards, the left arm will be teleoperated to approach and grasp on the other end. But due to the twisting axis alignment issues, the grasping pose needs to be adjusted and the left arm will move and adjust the other grasped end to make the deformable object reconstruct a cylinder for twisting. The adjustment can be achieved by realtime teleoperation with human operator's decisions and commands from the master device.
The selected experimental simulation states in the preparing stages on the slave side are shown in Fig.\ref{fig.robosim}.  

When the end effector of the teleoperated left arm was feasible to achieve an acceptable alignment tolerance which is manually defined by a cone with the cone angle of about $5^\circ$, i.e. $\delta_{tor} \approx 5^\circ$, the sequential object twisting manipulation can be further completed. 
During the coordinated twisting process, we tried to complete the object twisting task by assigning $\vartheta_R = -45^\circ$, $\vartheta_L = 45^\circ$, making $\vartheta_T = 90 ^\circ$. For the collision detection, we set the distance threshold to be the length of the object: $d_{thr} = \lvert P_1P_2 \rvert$.

\begin{figure}[htp] 
    \centering
    \vspace{-1mm}
    \includegraphics[width = 0.45\textwidth]{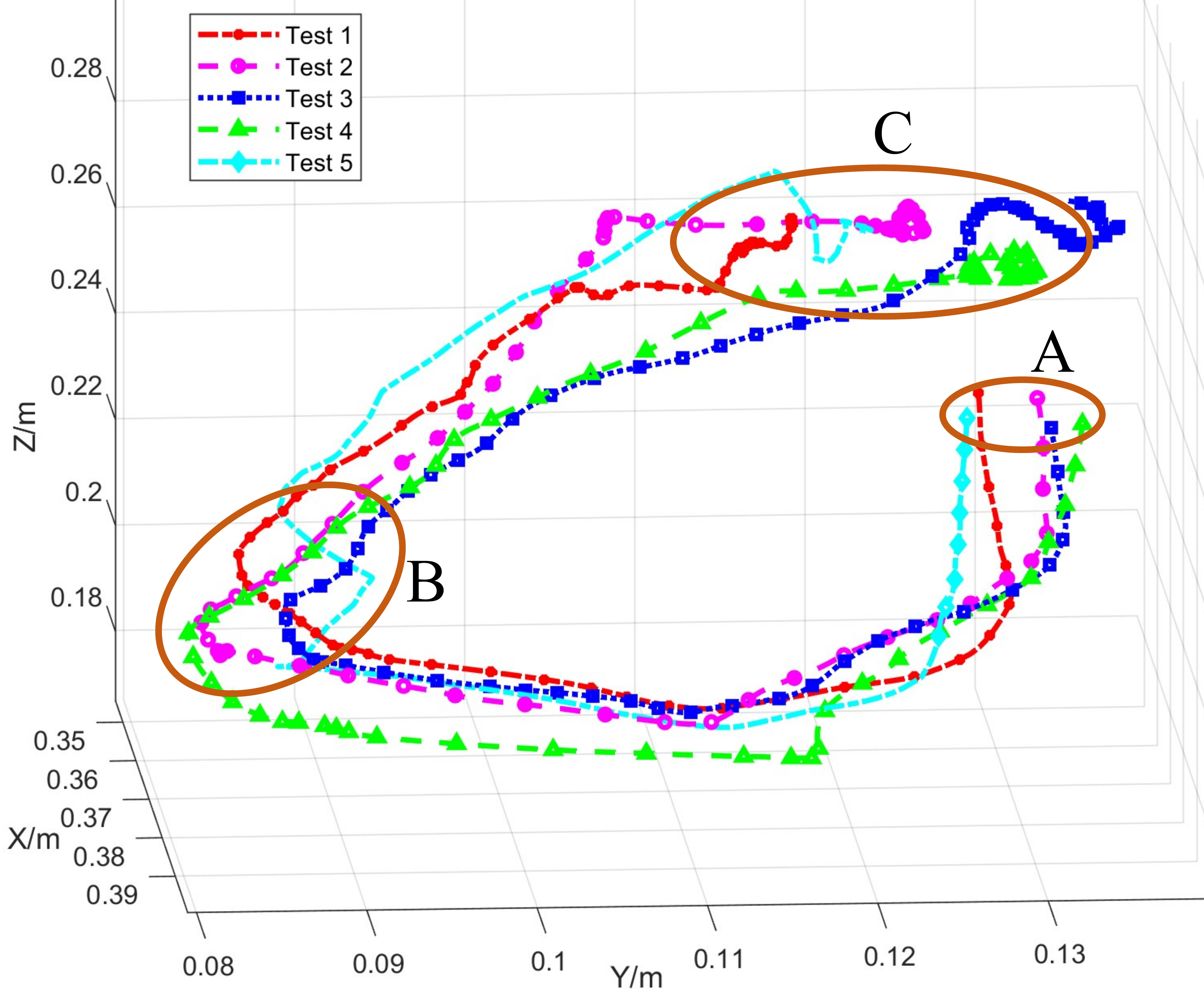}
    \caption{Left arm end-effector's teleoperating trajectories during the simulation tests of dual-arm coordinated twisting manipulation. The labelled area A: the starting points of the left arm's end-effector; B: the end-effector reaches the pendent end of the object and tries to grasp it with pose adjustment; C: the left arm moves the grasped end to this area to reconstruct the cylinder and rotate the object.}
    \label{fig.leftarmTeleTraj}
    \vspace{-4mm}
\end{figure}

\begin{figure}[h] 
    \centering
    \includegraphics[width = 0.50\textwidth]{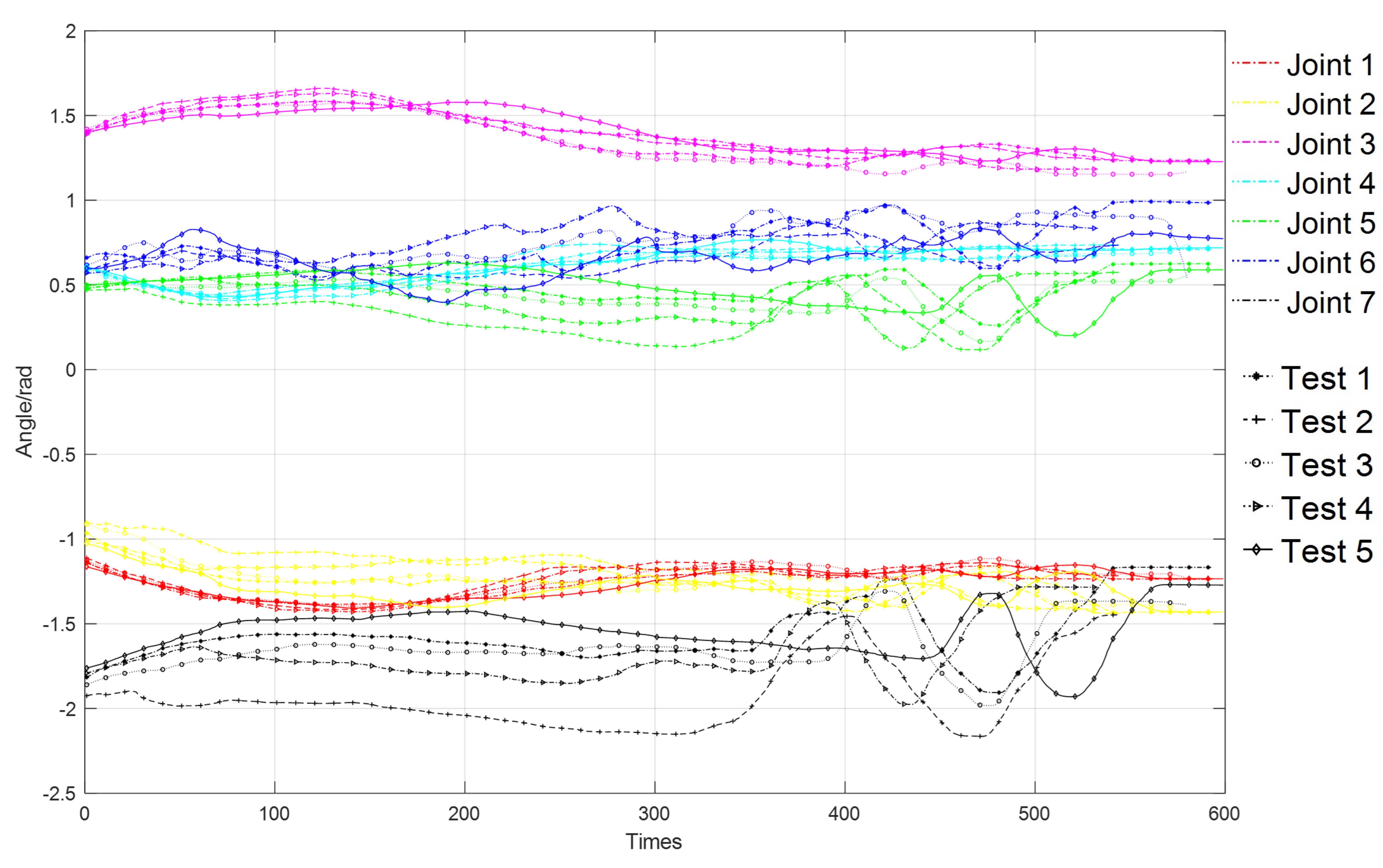}
    \caption{Joint state variation during the manipulation.}
    \label{fig.leftarmJnt}
    \vspace{-5mm}
\end{figure}

In this case, the right arm's end-effector will keep at the same position, and just rotate $-45^\circ$ about its $Z$ axis. While the left arm's end-effector will be teleoperated by human operator holding the master device in real-time. The pose adjustment and motion planning can be completed with the help of the human intelligence. To verify the feasibility of the proposed scheme for the dual-arm coordinated twisting manipulation, 5 simulation tests have been completed. The resulted trajectories of the left arm's end-effector are shown in Fig.\ref{fig.leftarmTeleTraj}. The variation of joint states during the process are presented in Fig.\ref{fig.leftarmJnt}. 

\section{
Discussion and 
Conclusion}

This paper mainly focuses on the common but challenging dual-arm coordinated manipulation for object twisting tasks. To complete the  task, teleoperation with human intelligence is introduced to control one of the arm, which would provide real-time decision for adjusting the feasible grasping pose, achieving twisting axis alignment even for deformable objects, and twisting the object with better coordination. With the consideration of self-collision detection, directional manipulability evaluation, and weighted configuration variation, the safety, efficiency, and stability of the dual-arm coordinated twisting manipulation is greatly enhanced.
The analysis and experimental simulation results briefly demonstrated the feasibility of the proposed method. Meanwhile, higher manipulability and smaller configuration variation for completing the twisting task can be achieved with the addressed constraints for dual arms. But the detailed quantitative characterization and contrastive analysis of our proposed manipulation approach are not provided here.
Thus, these detailed analysis and subsequent experimental studies on the real robot system will be presented in the near future.

\addtolength{\textheight}{-12cm}   


\balance


\bibliographystyle{IEEEtran}
\bibliography{IEEEabrv, mybibfile}


\end{document}